\documentclass[runningheads]{llncs}
\usepackage{cite}
\usepackage{graphicx}
\usepackage{epstopdf}
\graphicspath{{eps/}}
\usepackage{upgreek}
\DeclareGraphicsExtensions{.eps}
\usepackage[cmex10]{amsmath}
\usepackage{amsfonts}
\usepackage{amssymb}
\usepackage{enumerate}
\usepackage{url}
\newcommand{\keywords}[1]{\par\addvspace\baselineskip
\noindent\keywordname\enspace\ignorespaces#1}
\usepackage{multirow}
\usepackage{scrextend}
\usepackage{tabularx}
\usepackage{times}
\usepackage{fixltx2e}

\begin{document}
\title{Optimally Stabilized PET Image Denoising Using Trilateral Filtering}
\titlerunning{Optimally Stabilized PET Image Denoising Using Trilateral Filtering}
\author{Awais Mansoor
\and Ulas Bagci 
\thanks{Corresponding author: ulas.bagci@nih.gov. This research is supported by CIDI, the intramural research program of the National Institute of Allergy and Infectious Diseases (NIAID) and the National Institute of Biomedical Imaging and Bioengineering (NIBIB).}%
\and Daniel J. Mollura}
\institute{Department of Radiology and Imaging Sciences\\National Institutes of Health (NIH), Bethesda, MD 20892}
\authorrunning{A. Mansoor, U. Bagci, D. J. Mollura}
\maketitle

\begin{abstract}
Low-resolution and signal-dependent noise distribution in positron emission tomography (PET) images makes denoising process an inevitable step prior to qualitative and quantitative image analysis tasks. Conventional PET denoising methods either over-smooth small-sized structures due to resolution limitation or make incorrect assumptions about the noise characteristics. Therefore, clinically important quantitative information may be corrupted. To address these challenges, we introduced a novel approach to remove signal-dependent noise in the PET images where the noise distribution was considered as \emph{Poisson-Gaussian} mixed. Meanwhile, the generalized Anscombe's transformation (GAT) was used to stabilize varying nature of the PET noise. Other than noise stabilization, it is also desirable for the noise removal filter to preserve the boundaries of the structures while smoothing the noisy regions. Indeed, it is important to avoid significant loss of quantitative information such as standard uptake value (SUV)-based metrics as well as metabolic lesion volume. To satisfy all these properties, we extended bilateral filtering method into trilateral filtering through multiscaling and optimal Gaussianization process. The proposed method was tested on more than 50 PET-CT images from various patients having different cancers and achieved the superior performance compared to the widely used denoising techniques in the literature.
\end{abstract}

\section{Introduction}
Positron emission tomography (PET) is a 3-D non-invasive technique that uses radioactive tracers to extract physiological information. Like other low-photon counting applications, the reconstructed image in PET scanners has low signal-to-noise ratio (SNR), which can affect the diagnosis of disease through quantification of clinically relevant quantities such as standardized uptake value (SUV) and metabolic lesion volume. Therefore, a denoising mechanism for PET images has to be adopted as a preprocessing step for accurate quantification \cite{foster201476, sandouk2013318}.

Current approaches in PET denoising are mostly inherited from optical imaging where primary criteria for denoising is qualitative rather than quantitative. Among the effective methods derived from other biomedical imaging modalities, Gaussian smoothing \cite{chatziioannou1996290}, anisotropic diffusion \cite{demirkaya2002N271},  non-local means \cite{dutta2013e81390}, and bilateral filtering approaches \cite{hofheinz20111} either over-smooth the edges or violates the Poisson statistics of the data; hence, corrupting vital information. Recently, multiscale denoising approaches such as \cite{turkheimer2008657} and soft-thresholding methods \cite{bagci2013115} have been adapted for PET images to avoid over-smoothing of the edges. These methods have shown improvement in SNR compared to the conventional methods due to their superiority in preserving edges. However, multiscale methods do not perform well in the vicinity of weak boundaries because they fail to eliminate point-singularities. Soft thresholding approach, on the other hand, is promising and shown to be superior to others since the noise is modeled in more realistic way and boundaries of small-sized objects are preserved; however, optimal transformation of noise characteristics has not been addressed yet~\cite{bagci2013115}.

Parametric denoising methods in the literature consider the noise in PET images to be additive Gaussian~\cite{chatziioannou1996290}.  However, Gaussian assumption in PET images may result in the further loss of already poor resolution, increased blurring, and altered clinically relevant imaging markers. Recent attempts such as \cite{bagci2013115} used a more realistic Poisson-distributed noise assumption in PET images where authors first ``Gaussianize'' the Poisson measurement followed by unbiased risk estimation based denoising filtering. Gaussianization is achieved by applying a linear transformation such as a square-root and known as variance stabilizing transformation (VST)~\cite{anscombe1948246}. However,  the algebraic inverse VST used by this denoising method may be sub-optimal. Regarding these difficulties, we proposed a novel approach in this paper to denoise PET images using the optimal noise characteristics and a 3-D structure preserving noise removal filtering.

\section{Methods}
\label{sec:methods}
We consider the noise in PET images as a mixed distribution of Poisson and Gaussian. Our assumption stems from the \emph{Poisson} nature of photon-counting and \emph{Gaussian} nature of the reconstruction process. In our proposed methodology, a linear transformation (i.e., GAT) was first used to stabilize the noise variation optimally. Second, trilateral denoising filter (TDF) was developed and applied to the variance stabilized image. Finally, optimal exact unbiased inverse GAT (IGAT) was applied to obtain denoised PET images. 

\subsection{Generalized Anscombe Transformation (GAT)}
Let $x_i, i\in 1,\dots, N$ be the observed voxel intensity obtained through the PET acquisition system. Poisson-Gaussian noise distribution models each observation as an independent random variable $p_i$, sampled from Poisson distribution with mean $\lambda_i$, and scaled by a constant $\alpha>0$, and corrupted by Gaussian noise $\eta_i^*$ (with mean $\mu_i$ and variance $\sigma_i^2$) as
\begin{equation}
x^*_i=\alpha p_i+\eta_i^*,
\end{equation}
where $x_i^*\sim P\left( \lambda_i  \right)$ and $n_i^* \sim N\left( {\mu_i, \sigma^2_i} \right)$. \\
A variance stabilization transformation (GAT) assumes the existence of a function $f_\sigma$ that can approximately stabilize the variation of $x$ (i.e., ${\mathop{\rm var}} \left( {f\left( {{x^*_i}} \right)|{p_i}} \right) \approx \mbox{constant}$). Mathematically, for the Poisson-Gaussian noise model $x^*_i=\alpha p_i+\eta_i^*$, $f_\sigma(x)$ gives the optimal variance stabilization when $f_\sigma$ is piecewise linear and having the following form~\cite{starck1998image}:
\begin{equation}
f_\sigma(x) = \begin{cases} \frac{2}{\alpha}\sqrt {\alpha x + \frac{3}{8}\alpha^2 + {\sigma ^2}-\alpha\mu} &\mbox{if } x > -\frac{3}{8}-\sigma^2 \\
0 & \mbox{otherwise. } 
\end{cases}
\label{eq:GAT}
\end{equation}
Note that GAT equals to the traditional Anscombe transformation when noise is considered Poisson only ($\alpha=1$, $\sigma=0$, and $\mu=0$):
\begin{equation}
f_\sigma(x) = \begin{cases} 2\sqrt {x + \frac{3}{8} + {\sigma ^2}} &\mbox{if } x > -\frac{3}{8}-\sigma^2 \\
0 & \mbox{else}, 
\end{cases}
\label{eq:VST}
\end{equation}
where $x=\frac{x^*-\mu}{\alpha}$ and $\sigma=\frac{\sigma^*}{\alpha}$.

\subsection{Trilateral denoising filter (TDF)}
\label{sec:trilateral}
Trilateral denoising filter (TDF) is an extension to bilateral filter, and similar to the bilateral filters, TDF  belongs to an edge preserving Gaussian filtering family. Herein, we briefly describe the principal of TDF. 

Let the GAT transformed image $f_\sigma(x)$ be $f_G$.  A bilateral filter is an edge preserving filter defined as:
\begin{eqnarray}
{g_f}\left( i \right) = \frac{1}{{ k\left( i \right)}}\int { f_G\left( {i + \mathbf{a}} \right){w_1}\left( \mathbf{a} \right){w_2}\left( \mathbf{a} \right)\left( {\left\| {f_G\left( {i + \mathbf{a}} \right) -  f_G\left( i \right)} \right\|} \right)d\mathbf{a}}, \nonumber\\
 k\left( i \right) = \int {{w_1}\left( \mathbf{a} \right){w_2}\left( {\left\| { f_G\left( {i + \mathbf{a}} \right) - f_G\left( i \right)} \right\|} \right)d\mathbf{a}}, 
\end{eqnarray}
where $\mathbf{a}$ is an offset vector (i.e., defines a small neighborhood around the voxel $i$). The weight parameters $w_1$ and $w_2$ respectively measure the geometric and photometric similarities within a predefined local neighborhood $N_x$ and are designed as Gaussian kernels with standard deviations $\sigma_1$ (geometric range) and $\sigma_2$ (photometric range), the size of the neighborhood is adjusted using $\sigma_1$ and $\sigma_2$. Function $k\left( i \right)$ is the normalization factor.

A trilateral filter is a gradient preserving filter. It preserves the gradient by applying bilateral filter along the gradient plane. Let $\nabla f_G$ be the gradient of the GAT transformed image $f_G$, the trilateral filter is initiated by applying a bilateral filter on $\nabla f_G$,
\begin{eqnarray}
{g_f}\left( i \right) = \frac{1}{{\nabla k\left( i \right)}}\int {\nabla f_G\left( {i + \mathbf{a}} \right){w_1}\left( \mathbf{a} \right){w_2}\left( \mathbf{a} \right)\left( {\left\| {\nabla f_G\left( {i + \mathbf{a}} \right) - \nabla f_G\left( i \right)} \right\|} \right)d\mathbf{a}}, \nonumber\\
\nabla k\left( i \right) = \int {{w_1}\left( \mathbf{a} \right){w_2}\left( {\left\| {\nabla f_G\left( {i + \mathbf{a}} \right) - \nabla f_G\left( i \right)} \right\|} \right)d\mathbf{a}}. 
\end{eqnarray}
For refinement, subsequent second bilateral filter is applied using the $g_f$. Assuming $f_G\left( {i,a} \right) = f_G\left( {i + \mathbf{a}} \right) - f\left( x \right) - \mathbf{a}{g_f}\left( x \right)$ and let a neighborhood weighting function be
\begin{equation}
N_i = \begin{cases} 1 &\mbox{if } \left| {{g_f}\left( {i + \mathbf{a}} \right) - {g_f}\left( i \right)} \right| < c, \\
0 & \mbox{otherwise},
\end{cases}
\end{equation}
where $c$ specifies the size of adaptive region. Ultimately, the final trilateral smoothed image is given as
\begin{eqnarray}
{S_{TDF}}\left( f_G\left( i \right) \right) = f_G\left( i \right) + \frac{1}{{\nabla k\left( i \right)}}\int {\nabla f_G\left( {i,\mathbf{a}} \right){w_1}\left( \mathbf{a} \right){w_2}\left( {\nabla f_G\left( {i,\mathbf{a}} \right)} \right)N\left( {i,\mathbf{a}} \right)d\mathbf{a}}\nonumber \\
\nabla k\left( i \right) = \int {{w_1}\left( \mathbf{a} \right){w_2}\left( {\nabla f_G\left( {i,\mathbf{a}} \right)} \right)} N\left( {i,\mathbf{a}} \right)d\mathbf{a}\mbox{.  }
\end{eqnarray}
For TDF, $\sigma_1$ (geometric range) is the only input parameter to trilateral filter, $\sigma_2$ (photometric range) can be defined as (see \cite{wong2004820} for the justifications)
\begin{equation}
{\sigma _2} = 0.15\left| {\mathop {\max }\limits_{i} {{\overline g }_f}\left( i \right) - \mathop {\min }\limits_{i } {{\overline g }_f}\left( i \right)} \right|
\end{equation}
and ${{{\overline g }_f}\left( i \right)}$ is the mean gradient of the GAT transformed image.

\subsection{Optimal exact-unbiased inverse of generalized Anscombe transformation (IGAT)}
After obtaining the $f_{{\sigma}}\left( x \right)$, we can treat the denoised data $\mathcal{D}={S_{TDF}}\left( f_G \right)$ as the expected value $E\left\{ {{f_{{\sigma}}}\left( x \right)|\lambda,{\sigma}} \right\} $. The closed form of optimal exact unbiased IGAT is given as, 
\begin{eqnarray}
\mathcal{I} = E\left\{ {{f_{{\sigma}}}\left( x \right)|\lambda,{\sigma}} \right\} &=& \int\limits_{ - \infty }^{ + \infty } {{f_{{\sigma}}}\left( x \right)p\left( {x|\lambda,{\sigma}} \right)dx}\nonumber\\
&=& \int\limits_{ - \infty }^{ + \infty } {2\sqrt {x + \frac{3}{8} + {\sigma ^2}} \sum\limits_{k = 0}^{ + \infty } {\left( {\frac{{{\lambda^k}{e^{ - \lambda}}}}{{k!\sqrt {2\pi {\sigma ^2}} }}{e^{ - \frac{{{{\left( {x - k} \right)}^2}}}{{2{\sigma ^2}}}}}} \right)dx.} }
\end{eqnarray}
The optimal inverse $\mathcal{I}$ (in maximum likelihood sense) is applied to the denoised data $\mathcal{D}$ followed by scaling and translation (i.e., $\alpha\mathcal{I}(\mathcal{D})+\mu$) for obtaining the denoised PET image.

\section{Experiments and Results}
We performed a comprehensive analysis and comparison of our approach with widely used denoising methods (i) Gaussian filter, (ii) bilateral filter \cite{hofheinz20111}, (iii) anisotropic diffusion filter \cite{demirkaya2002N271}, and (iv) our presented trilateral filter; without noise stabilization, with Poisson noise stabilization (VST, eq. \ref{eq:VST}), and Poisson-Gaussian noise stabilization (GAT, eq. \ref{eq:GAT}).\\ 
\textbf{\underline{Data}: }We used both phantoms as well as clinical data for evaluation of the denoising algorithms.\\
\textbf{\underline{Phantoms}: }Data for the SNM Germanium Phantom were acquired using a GE DSTE-16 PET/CT scanner (16-row MDCT)~\cite{petphantom}. A total of 6 scans were acquired consisting of 5, 7, and 30 minute acquisitions using both 3D (no septa) and 2D (with septa) modes. A total of 36 reconstructions were completed consisting of 3 OSEM and 3 filtered back projection (FBP) reconstructions per scan. Both OSEM and FBP images were reconstructed using smoothing filters of 7, 10, and 13 mm (Fig. \ref{fig:line}). The resulting activity concentrations were also converted into target-to-background (T/B) ratios, and SUVs.

\noindent\textbf{\underline{Clinical Data}: }With the IRB approval, PET-CT scans from 51 patients pertaining to various cancer diseases were collected retrospectively. All patients underwent PET-CT imaging (on Siemens Biograph 128 scanners) such that patients were instructed to fast for a minimum of 6-hours before scanning. At the end of the 6 hour period, 3.7-16.3 mCi (median=10.1 mCi, mean=9.45 mCi) of $^{18}$F-FDG was administered intravenously to the patients depending on the body weight. PET images were obtained in two dimensional mode. The intrinsic spatial resolution of the system was 678.80 mm. CT was performed primarily for attenuation correction with the following parameters: section thickness, 3 mm; tube voltage, 120 kVp; tube current, 26 mAs; field of view, $500\times500$ mm.

\subsection{Qualitative and Quantitative Evaluations}
\begin{figure*}[htp]
\centering
\includegraphics[scale =0.22]{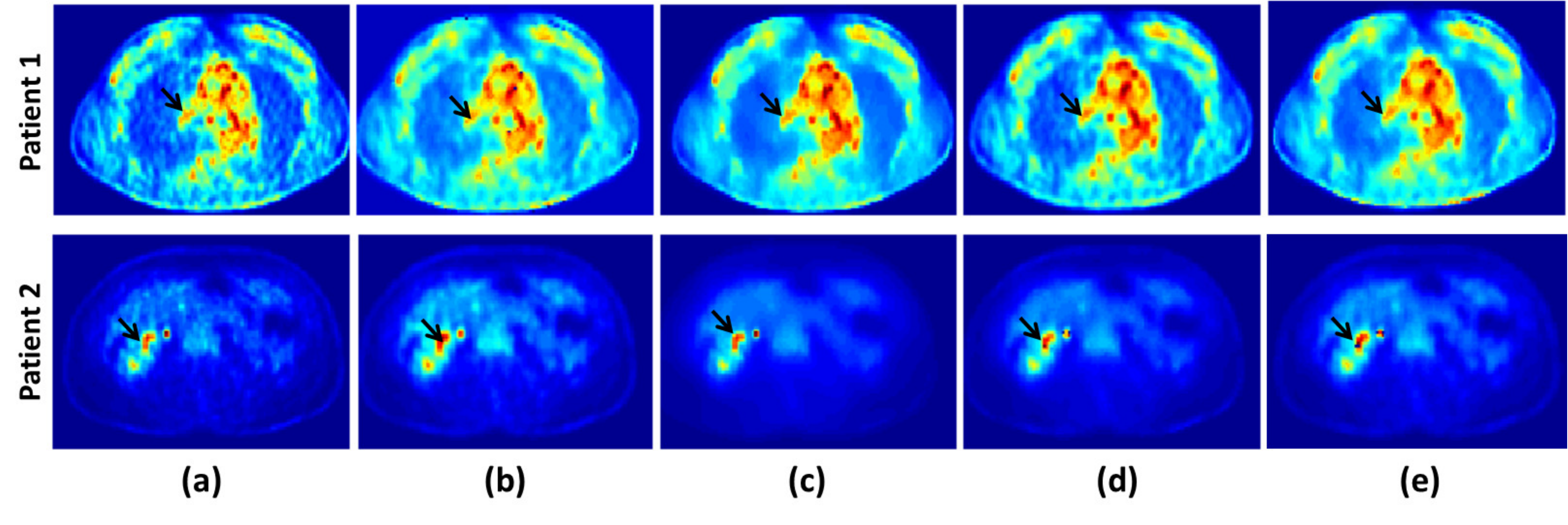}  	
\caption{\small{Qualitative evaluation of the proposed method with current methods. Each row shows PET scans from different subjects. Each row shows different patient. (a) original noisy image, (b) GAT+anisotropic filter, (c) GAT+bilateral filter, (d) VST+trilateral filter, and (e) GAT+trilateral filter (proposed). \emph{Black} arrows indicate the object of interest where edge information is preserved.}}
  \label{fig:results}
\end{figure*}

We qualitatively performed a comparison of our proposed method with above mentioned methods with different combinations of variance stabilization (no stabilization, VST, GAT). Results of GAT+anisotropic filter, GAT+bilateral filter, VST+TDF, and GAT+TDF are presented in Fig. \ref{fig:results}. As pointed out with arrows, the boundary contrast is the highest in the proposed GAT+TDF; whereas other methods either over-smoothed or over-saturated the noisy areas. This is ensured using the TDF by employing an iterative approach coupled with narrow spatial window to preserve edges at finer scales. Also VST+TDF result (Fig. \ref{fig:results}) verifies that Poisson noise assumption is more realistic than Gaussian but suboptimal with respect to the Poisson-Gaussian assumption.

\begin{figure*}[htp]
  \centering
	\includegraphics[scale =0.17]{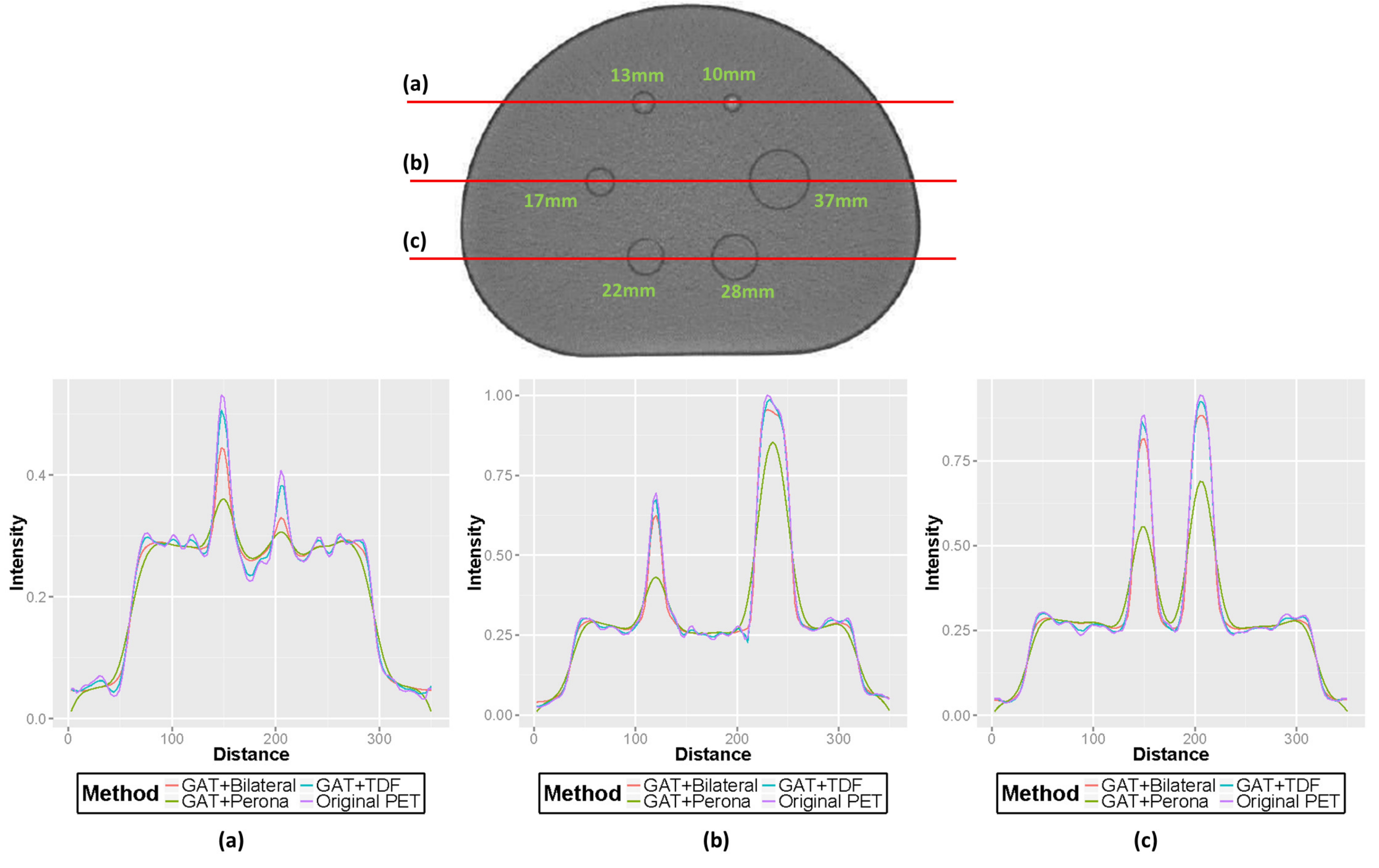}	
  \caption{\small{Profile plots on all six spheres for simulated phantom dataset.}}
  \label{fig:line}
\end{figure*}

To evaluate the potential loss of resolution and enhanced blurring after the denoising procedure, we employed \emph{line profiling} through lesion ROIs in phantom image (shown in Fig. \ref{fig:line}). Superiority of GAT+TDF can be readily depicted from the figure. 

\begin{figure}[htp]
\centering
{\includegraphics[scale =0.315]{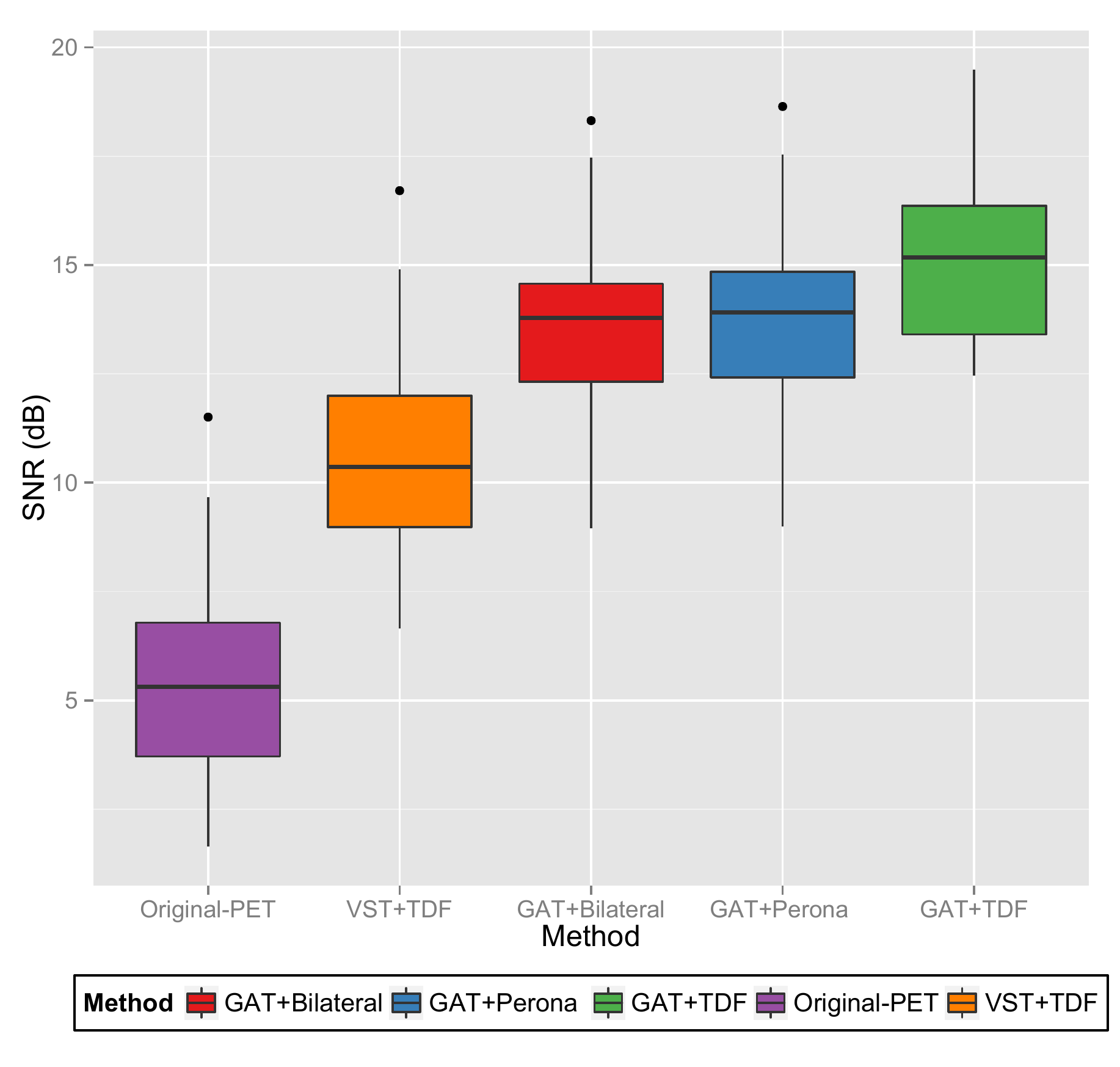}}
{\includegraphics[scale =0.315]{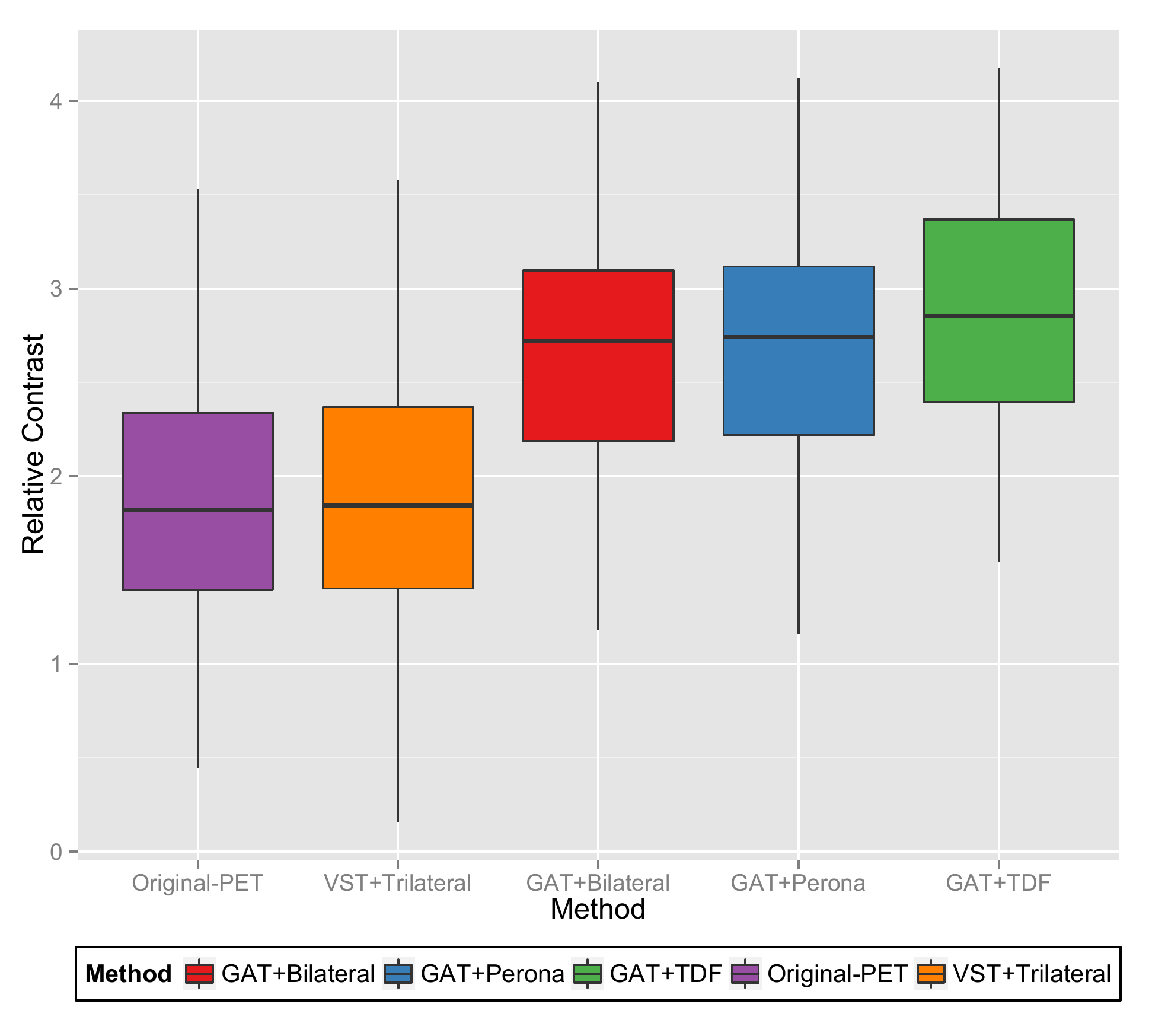}}
\caption{\small{Box-plots for (a) SNR (eq. \ref{eq:snr}) and (b) relative contrast (eq. \ref{eq:rc}) are demonstrated.}}
\label{fig:boxplots}
\end{figure}

For quantitative evaluation of PET imaging markers, we manually drew the region of interest (ROI) around lesions/tumors and large homogeneous regions such as liver and lung in the PET scans. Quantitative information including SNR was then extracted from these ROIs as shown in the boxplot (Fig. \ref{fig:boxplots}(a)). The SNR of the image from selected ROIs was defined as
\begin{equation} 
SN{R_i} = 20{\log _{10}}\left( {\frac{{{m_i}}}{{{\sigma _i}}}} \right),
\label{eq:snr}
\end{equation}
where $m_i$ is the mean and $\sigma_i$ is the variance of the $i^{th}$ ROI. In addition, the relative contrast (RC) of the ROIs (Fig. \ref{fig:boxplots}(b)) was calculated using the following relationship \cite{bagci2013115}
\begin{equation}
R{C_i} = \frac{{\left| {{m_i} - {M_B}} \right|}}{{\sqrt {{\sigma _i}{\sigma _B}} }}
\label{eq:rc}
\end{equation}

In clinics, an optimal denoising method is expected to reduce the noise and increase the SNR whilst preserving the clinically significant information such as SUV\textsubscript{max}, SUV\textsubscript{mean}, metabolic tumor volume, etc. To assess how these values were affected from the denoising process, we measured significance of the percentage change in SUV\textsubscript{max} and SUV\textsubscript{mean} in different ROIs using Kruskal-Wallis test, a non-parametric \emph{one-way analysis of variance}. The results of the test for SUV\textsubscript{mean} and SUV\textsubscript{max} together for our method in comparison other methods is presented in Table \ref{table:wallis}. As shown in the table, the change in SUV matrices are not statistically significant with our approach. Other imaging markers (SUV\textsubscript{max} and metabolic tumor volume) have shown similar trends.
\begin{center}
\begin{table}
\centering
\caption{Kruskal�-Wallis one-way analysis of variance for different denoising methods of SUV\textsubscript{mean} and SUV\textsubscript{max} together. df=Degrees-of-freedom.}
\begin{minipage}[t]{0.45\linewidth}
\begin{tabular}{p{3cm} c c c c c}
\hline
\textbf{Method} &\textbf{df}  & $\mathbf{\chi^2}$ & \textbf{$p$-value}\\
\hline
Gaussian filter&1&18.86&0.001\\
Bilateral filter&1&3.77&0.05\\
Perona-Malik&1&3.67&0.05\\
TDF&1&3.24&0.07\\
VST+Gaussian filter&1&11.32&0.001\\
VST+Perona-Malik&1&2.75&0.09\\
\hline
 \end{tabular}
\end{minipage}
\begin{minipage}[t]{0.45\linewidth}
\begin{tabular}{p{3cm} c c c c c }
\hline
\textbf{Method} & \textbf{df}  & $\mathbf{\chi^2}$ & \textbf{$p$-value}\\
\hline
VST+TDF&1&1.12&0.289\\
GAT+Gaussian filter&1&2.34&0.12\\
GAT+Bilateral filter&1&0.27&0.60\\
GAT+Perona-Malik&1&0.28&0.59\\
\textbf{GAT+TDF (proposed)}&1&0.18&0.67\\
\\
\hline
 \end{tabular}
\end{minipage}
\label{table:wallis}
\end{table}
\end{center}

\section{Discussion and Conclusion}
Inspired by the study \cite{bagci2013115}, in which the authors showed that variance stabilization transformation  is an important step in denoising, we proposed a novel approach for denoising PET images. In particular, we presented an optimal formulation for variance stabilization transformation and its inverse. Furthermore, a more  realistic noise distribution of PET images (i.e., Poisson-Gaussian) was considered. For smoothing of PET images after Gaussianizing the noise characteristics, we extended bilateral filtering into trilateral denoising filter that is able to preserve the edges as well as quantitative information such as SUV\textsubscript{max} and SUV\textsubscript{mean}. Experimental results demonstrated that our proposed method: (i) effectively eliminate the noise in PET images, (ii) preserve the edge and structural information, and (iii) retain clinically relevant details. As an extension to our approach, we plan to integrate our algorithm with partial volume correction step in order to study the impact of the combined method on object segmentation. We are also determined to compare our algorithm with the trending soft-thresholding and non-local means based algorithms in a larger evaluation platform where objective comparison and assessment of the denoising steps will be possible.

\bibliographystyle{splncs}

\end{document}